\documentclass{article}

\usepackage{microtype}
\usepackage{graphicx}
\usepackage{subcaption}
\usepackage{booktabs} 

\usepackage{hyperref}

\usepackage[accepted]{icml2023}

\usepackage{amsmath}
\usepackage{amssymb}
\usepackage{mathtools}
\usepackage{amsthm}

\usepackage[capitalize,noabbrev]{cleveref}

\theoremstyle{plain}

\theoremstyle{definition}

\theoremstyle{remark}

\newtheorem{example}{Example}

\usepackage{booktabs}
\usepackage{float}
\usepackage{inconsolata}

\icmltitlerunning{Assessing GPT4-V on Structured Reasoning Tasks}

\begin{document}

\twocolumn[
\icmltitle{Assessing GPT4-V on Structured Reasoning Tasks}



\icmlsetsymbol{equal}{*}

\begin{icmlauthorlist}
\icmlauthor{Mukul Singh}{india}
\icmlauthor{Jose Cambronero}{us}
\icmlauthor{Sumit Gulwani}{us}
\icmlauthor{Vu Le}{us}
\icmlauthor{Gust Verbruggen}{belgium}

\end{icmlauthorlist}

\icmlaffiliation{india}{Microsoft, India}
\icmlaffiliation{belgium}{Microsoft, Belgium}
\icmlaffiliation{us}{Microsoft, United States}

\icmlcorrespondingauthor{Mukul Singh}{singhmukul@microsoft.com}

\icmlkeywords{Machine Learning, ICML}

\vskip 0.3in
]




\begin{abstract}
Multi-modality promises to unlock further uses for large language models.
Recently, the state-of-the-art language model GPT-4 was enhanced with vision capabilities.
We carry out a prompting evaluation of GPT-4V and five other baselines on structured reasoning tasks, such as mathematical reasoning, visual data analysis, and code generation.
We show that visual Chain-of-Thought, an extension of Chain-of-Thought to multi-modal LLMs, yields significant improvements
over the vanilla model.
We also present a categorized analysis of scenarios where these models perform well and where they struggle, highlighting challenges associated with coherent multimodal reasoning.
\end{abstract}

\section{Introduction}

Visual instruction tuning \cite{liu2023visual} allows language models to answer questions about images.
By leveraging a dataset of captioned images, strong language models (like GPT-4) can be used to generate high-quality training data that spans conversation, detailed descriptions and reasoning about concepts found in these images.
The visual instruction tuned model is then able to reason (to some extent) about the input image \cite{zhang2023multimodal}.
OpenAI GPT-4V is the state-of-the-art language model GPT-4 \cite{OpenAI2023GPT4TR} augmented with a vision component.

\begin{example}
    The following image and part of the associated complex reasoning instruction are part of the LLaVA dataset \cite{liu2023visual}. The image and \textbf{Q} are provided to the model, and it is trained to return \textbf{A}.\\[.5em]
    \begin{minipage}{.49\columnwidth}
    \includegraphics[width=\textwidth]{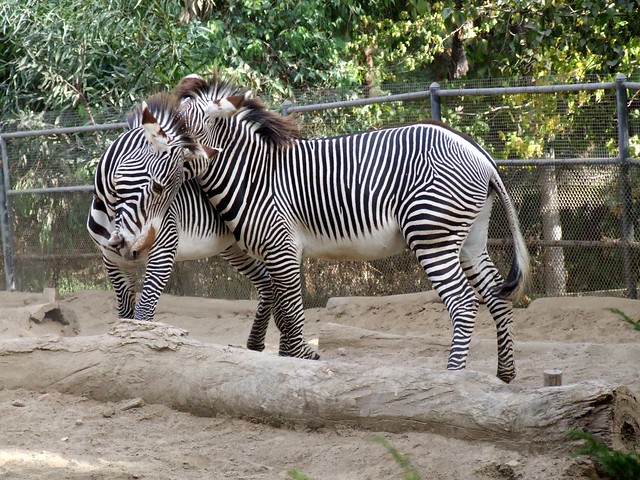}
    \end{minipage}%
    \hfill
    \begin{minipage}{.49\columnwidth}
        \small
        \textbf{Q:} What can be inferred about the relationship between these zebras?\\
        \textbf{A:} The zebras are standing near each other and nuzzling their heads together behind the fence in their enclosure. This behavior suggests a bonding moment or a sign of affection between them.
    \end{minipage}
\end{example}

In large language models, such as GPT-4, explicitly eliciting reasoning significantly improves performance.
A popular method is Chain-of-Thought (CoT) where a literal instruction to ``think step by step'' causes the model to first generate some thoughts, and then use these thoughts to generate a final answer \cite{wei2022chain}.
Similar to complex reasoning in  LLaVA, multi-modal Chain-of-Thought (m-CoT) \cite{zhang2023multimodal} leverages generating CoT data to fine-tune a multi-modal model to illicit reasoning about and answer scientific questions over graphics.

In this work, we explore the extent to which GPT-4V can \emph{reason} in different domains, such as code and math, when part of the inputs are images or can be represented as images.
More specifically, we evaluate GPT-4V in four domains.

\begin{enumerate}
    \item Mathematical questions with visual context from the MathVista dataset~\cite{lu2023mathvista}.
    \item Questions about data charts from the ChartQA dataset~\cite{masry-etal-2022-chartqa}.
    \item Abstraction and reasoning problems over objects in grids from the ARC dataset~\cite{chollet2019measure,arc}.
    \item Generating SQL from NL given a rendered table from the Spider dataset~\cite{yu-etal-2018-spider}.
\end{enumerate}
Each of these domains requires specific reasoning that might not be present in the training data.

Rather than fine-tuning a model, like m-CoT, we adapt the classical CoT approach to multi-modal language models, and show that it significantly outperforms the base (visual instruction tuned) model.
Besides using the exact reasoning structure from m-CoT, we introduce an improved, three-step reasoning process to further improve performance, called \emph{visual Chain-of-Thought} (v-CoT).
In v-CoT, we instruct the model to (1) extract relevant information about the image, (2) use this relevant information to reason about the problem, and (3) state the final answer.
Note that step (1) is not the same as a regular image captioning step, as it is also conditioned on the associated question along with the image.

\begin{figure}[tb]
    \centering
    \includegraphics[width=\columnwidth]{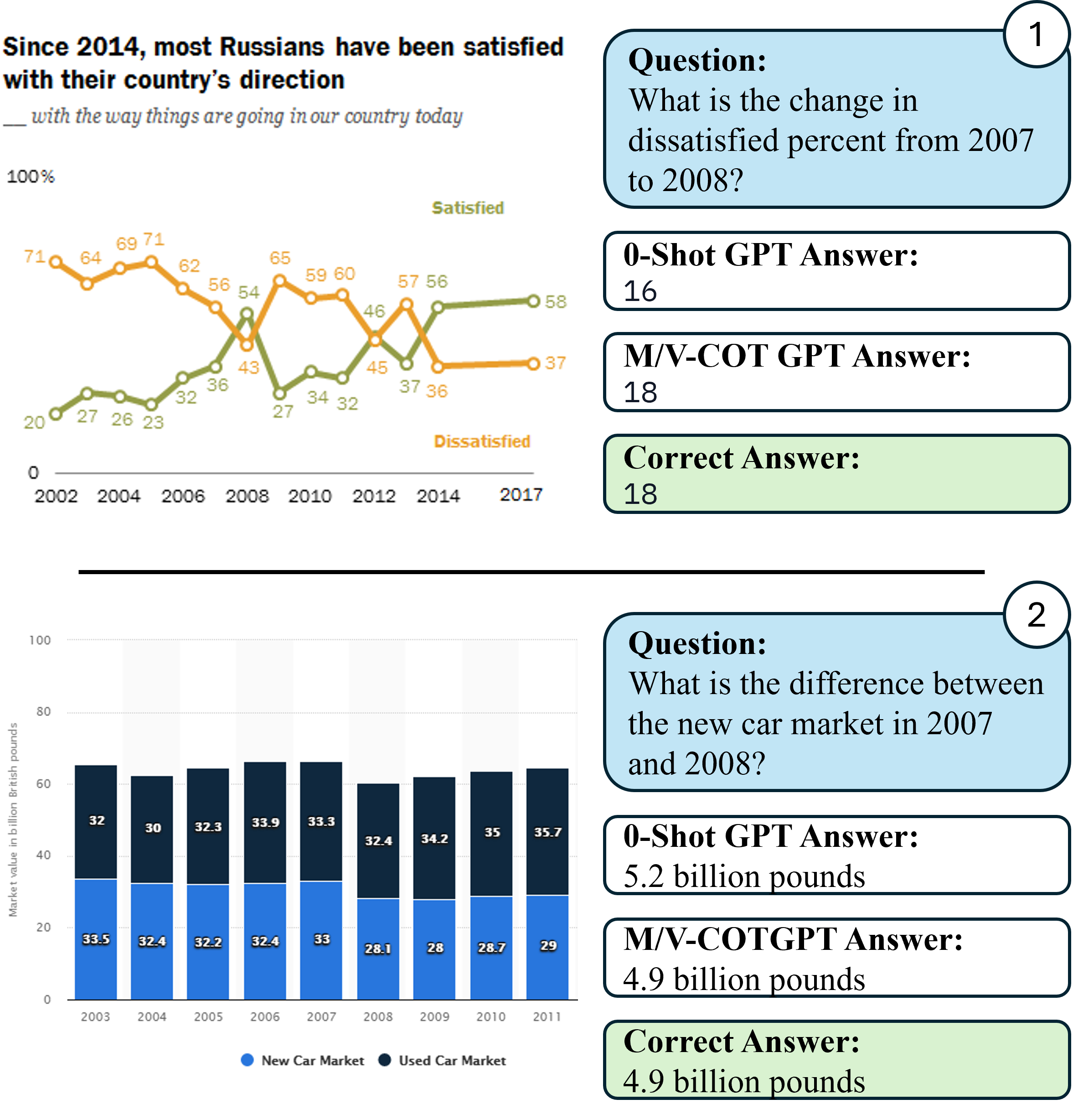}
    \caption{Without reasoning, the model does not find the correct answer. Both m-CoT and v-CoT elicit sufficient reasoning.}
    \label{fig:reasoning-example}
\end{figure}

\begin{example}
Figure~\ref{fig:reasoning-example} shows two examples where without reasoning, the model is not able find the correct answer.
\end{example}

\begin{figure}[htb]
\includegraphics[width=\columnwidth]{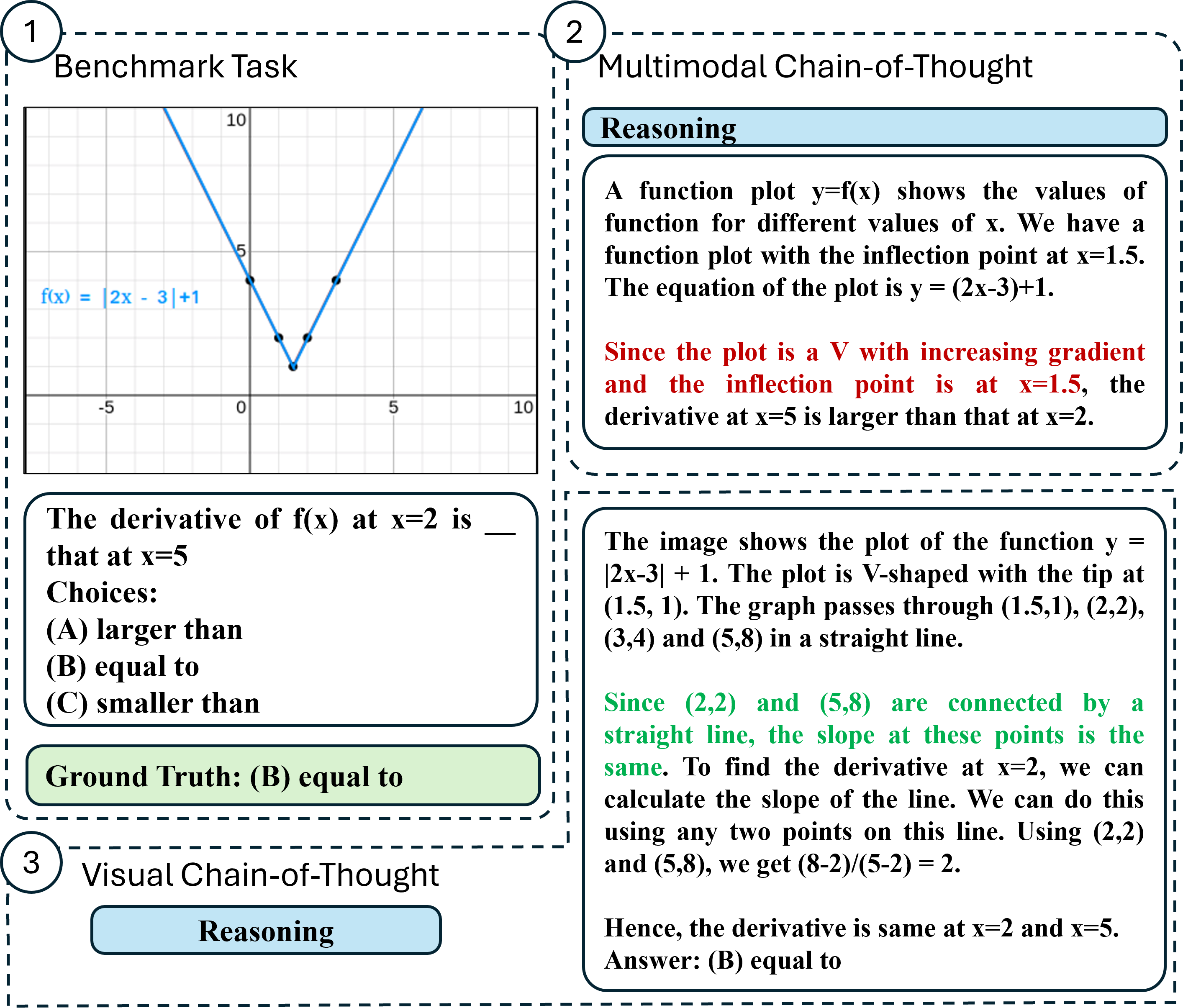}
\caption{
(1) An example task, (2) the output generated m-CoT, and (3) the output generated with v-CoT.
Red and green text highlight incorrect and correct reasoning, respectively.}
\label{fig:vcot-intro}
\end{figure}

\begin{example}
Figure~\ref{fig:vcot-intro} shows an example of reasoning with m-CoT and v-CoT.
By first describing relevant information, the model is then able to reason about that information to obtain the right result.
\end{example}

Our results show that for MathVista, ChartQA and Spider, GPT-4V outperforms baselines, including CoT and Program-of-Thought (PoT) prompting with GPT-4 over captions generated by InstructBlip \cite{dai2023instructblip}.
For ARC, we find that directly prompting over the vision component actually results in lower performance than captioning and CoT or PoT.
We find that using v-CoT improves 1.5 -- 9.3 percentage points over vanilla GPT-4V, and 1.1 -- 4.7 percentage points compared with the m-CoT prompt.

In summary, we make the following contributions:
\begin{itemize}
\item A comparative evaluation of GPT-4V and (captioning + GPT-4) on structured reasoning tasks.
\item Visual Chain-of-Thought (v-CoT) as an extension of CoT to multi-modal LLMs that first asks the model to extract relevant properties to reason over.
      v-CoT improves on vanilla GPT-4V by 1.5 -- 9.3 percentage points.
\item A thorough analysis of recurring patterns to understand GPT-4V's performance.
\end{itemize}

\section{Related Work}

Prior work has explored combining
multiple modalities using transformer-based models, including
visual and text \cite{
hu2023bliva,
dai2023instructblip,
li2023blip2,
lin2023sphinx}, speech
and text \cite{zhang2023speechgpt},
video and 
audio \cite{
zhang2023video,maaz2023video}, and the union of these modalities \cite{wu2023next}. 
In addition, there is a long history of
models taking inputs in one modality (like image or
text) and generating outputs in another (like text or images, respectively) \cite{
hossain2019comprehensive,
reed2016generative}.
For the image-text multimodal models, visual question-answering (VQA) \cite{antol2015vqa, agrawal2016vqa} and conditioned captioning \cite{li2023blip2, dai2023instructblip} are the most popular tasks where the input is an image and optionally some text and the output is text conditioned on botht he input image and text.
In this work, we present an empirical evaluation
of an existing 
state-of-the-art vision-text multimodal model (GPT-4V) on 
structured tasks like mathematical reasoning
and code generation, and consider a setting similar to VQA where both images and text are used as input and the target output is text.

To evaluate these multi-modal models a large variety of benchmarks have been introduced, including VQA \cite{antol2015vqa} and RefCOCO \cite{yu2016modeling}.
In domains of structured reasoning, recent datasets include MathVista~\cite{lu2023mathvista},
which collects a large number of mathematical reasoning tasks, and ChartQA~\cite{masry-etal-2022-chartqa}, which presents questions to be answered based on data analysis plots.
Recently, MMMU~\cite{yue2023mmmu} introduced college-level multi-modal tasks across a large range of domains including technical areas.
Our evaluation also considers rendering existing structured inputs (tables) as images to perform text-to-SQL generation over them, using the popular Spider dataset~\cite{yu-etal-2018-spider}.

Chain-of-Thought \cite{wei2022chain} reasoning traces were used to fine-tune a multi-modal CoT model \citet{zhang2023multimodal}.
We take inspiration from m-CoT in our prompting experiments with GPT-4V.
Leveraging a larger model (GPT-4), our visual guided prompting  techniques are aimed at generalizing better to new modes of reasoning.
\section{Evaluation Setup}

First, we describe v-CoT and all baselines used in our evaluation.
Second, we describe the benchmarks on which these approaches are evaluated.

\subsection{Methods}

We divide our baselines in three categories: multi-modal prompting strategies (m-CoT and v-CoT), directly using large instruction-tuned vision-text models, and captioning with instruction-tuned models plus reasoning over these captions with GPT-4.

\subsubsection{m-CoT and v-CoT}
\begin{figure}
\centering
\includegraphics[width=0.8\linewidth]{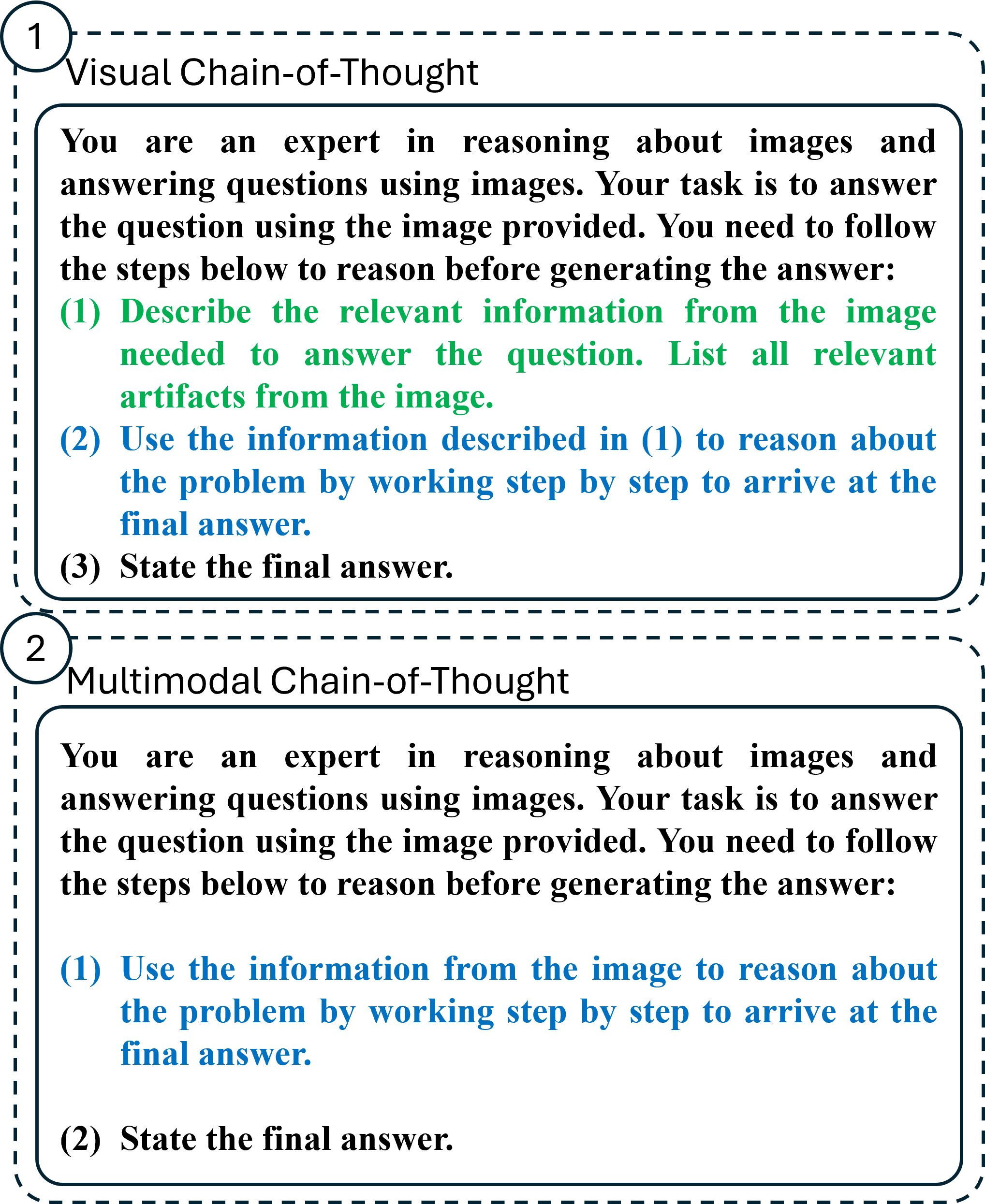}
\caption{Prompt structure for v-CoT (1) and m-CoT (2) prompts.
Blue highlights the shared m-CoT instruction and green highlights our extension.
}
\label{fig:vcot-approach}
\end{figure}

Figure~\ref{fig:vcot-approach} compares the corresponding prompts for v-CoT and m-CoT.
The latter works as traditional CoT but over both text and image inputs.
Specifically, the instruction to perform reasoning specifies that the image can be used to arrive at final answer in a step-by-step fashion.

The v-CoT prompt makes two small changes: (1) it instructs to describe relevant information and relevant image artifacts required to answer the question, and then (2) it asks to reason about this information to obtain the final answer.
These artifacts act like predicates that can be used to reason over the image to solve the task, and can range from concrete values in the image (\emph{the maximum value of the red line is 23}) to abstract concepts (\emph{a right angled triangle}).
Figure~\ref{fig:vcot-intro} shows an example where the model first identifies the points the line passes through even when this is not explicitly stated in the image.

\subsubsection{Instruction tuning}

\paragraph{Sphinx} \cite{lin2023sphinx} is a large multi-modal model (LMM) with
    multi-purpose visual instruction-following capabilities
    Sphinx has been trained on a variety of vision and language alignment tasks like object detection, visual question answering and region level captioning and achieves state-of-the-art performance on these.

\paragraph{Blip2} \cite{li2023blip2} is a text and image pre-training technique that bootstraps vision-language pre-training from off-the-shelf frozen pre-trained image encoders and frozen large language models.
The model combines existing vision and language models and pre-trains them on language and vision alignment tasks. Blip2 has been trained on various image-language tasks like conditional image captioning.

\paragraph{InstructBlip} \cite{dai2023instructblip} is an extension that applies instruction tuning to Blip-2 for language and vision tasks like visual question answering.

\subsubsection{Captioning + Prompting}

\textbf{InstructBlip + GPT4} with \textbf{Chain-of-thought (CoT)} \cite{wei2022chain} and
    \textbf{Program-of-thought (PoT)} \cite{chen2023program} prompting of
    GPT-4.
CoT and PoT have shown state-of-the-art performance in text and code generation tasks.
To incorporate images, we first generate a caption for the image using InstructBlip, add the caption to the input prompt for GPT-4, and provide instructions for CoT or PoT.

\subsection{Benchmarks}

This section describes the benchmark tasks used in our evaluation.
For all datasets, we perform exact match checks compared to the ground truth.
We manually inspect failures to ensure they are not matching issues.

\begin{figure*}
    \centering
    \includegraphics[width=\textwidth]{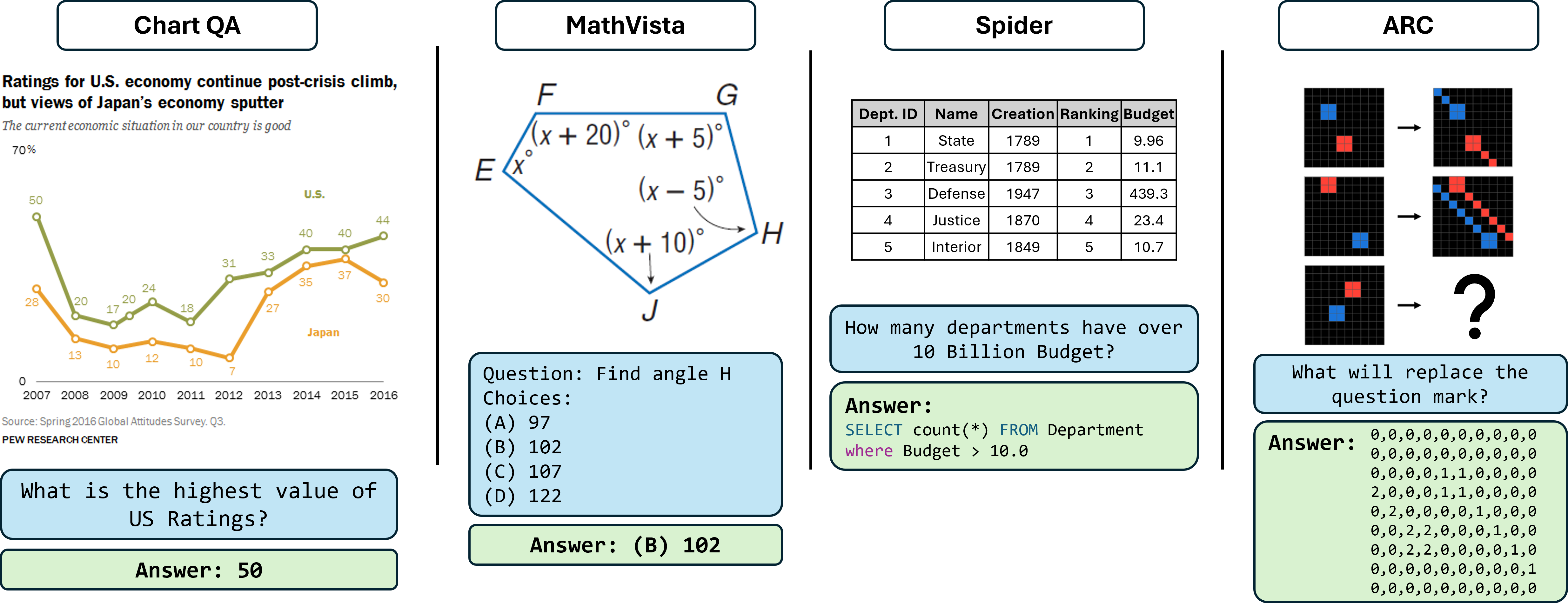}
    \caption{Sample tasks from each benchmark dataset. We show the image and the associated text prompt for the dataset along with the correct answer for the task. For Chart QA and MathVista the answer is a choice or numeric value; For Spider the answer is the correct SQL query; For ARC the output is the correct pixel matrix ($0 \rightarrow black; 1 \rightarrow red; 2 \rightarrow blue$).
    }
    \label{fig:dataset-samples}
\end{figure*}

Table~\ref{table:summary-benchmarks} summarizes our four benchmarks, and 
Figure~\ref{fig:dataset-samples} presents an example task from benchmark.
To mitigate computational costs, we sample 20\% tasks uniformly at random from each dataset.

\begin{table}
\centering
\small
\begin{tabular}{lllll} 
\toprule
\textbf{Dataset} & \textbf{Tasks} & \textbf{Image} & \textbf{Task} \\ \midrule
ChartQA (test) & 302 & Chart & VQA \\
MathVista (testmini) & 200 & Diagram & VQA \\
ARC (test) & 80 & Blocks & Pattern \\
Spider (test) & 206 & Table & Code generation \\
\bottomrule
\end{tabular}
\caption{Tasks from our evaluation benchmarks,
we present the total number of tasks,
image description,
and task description.
We sample 20\% from each dataset to mitigate
computational costs.}
\label{table:summary-benchmarks}
\end{table}

\subsubsection{Mathematical Reasoning}
We evaluate mathematical reasoning given a visual context with the recently introduced MathVista dataset \cite{lu2023mathvista}.
These problems require reasoning over images which may contain diagrams, tables or other images. Each task
consists of an image and question
pair.
Questions in MathVista can be multiple-choice, free-form with single value as result, or free-form with a list as a result.
We sampled tasks from the testmini split of MathVista, which resulted in multiple choice and single-value free-form questions\footnote{testmini has only 2 free-form list questions}.

\subsubsection{Visual Data Analysis}
We evaluate the ability to answer questions based on charted data---an important
skill for visual data analysis---using the ChartQA dataset \cite{masry-etal-2022-chartqa}.
Each tasks consists of a plot as a rendered image and a question about this plot.
These questions are short and objective, with 2-3 word or numeric value answers.
An example question is ``\textit{How many crimes were committed in 2020?}''
The question is passed in the prompt, preceded by the following instruction:

\texttt{Answer the question using the image. Only give the exact answer in 2-3 words or as a numeric value.}

\subsubsection{Visual Abstraction and Extrapolation}
We evaluate the ability of multi-modal models to solve program-synthesis-like tasks over grids that require abstraction and extrapolation by using the ARC dataset \cite{arc}.
Each ARC task consists of a set of examples, where an example consists of an input and output grid, and a new input---the model should predict the associated output grid.
Grids are represented as comma-separated-value (CSV) table of integers.
We use the following instruction

\texttt{Generate the transformed representation that will replace the question mark by looking at the example figure transformation. Generate as a <gridsize> Grid where 0 denotes black, 1 denotes red and 2 denotes blue.}

where we replace \texttt{<gridsize>}  with the corresponding dimensions specified in the individual benchmark task.

\begin{figure*}[t]
    \captionsetup{justification=centering}
    \centering
    \begin{subfigure}{0.49\textwidth}
        \centering
        \includegraphics[width=\textwidth]{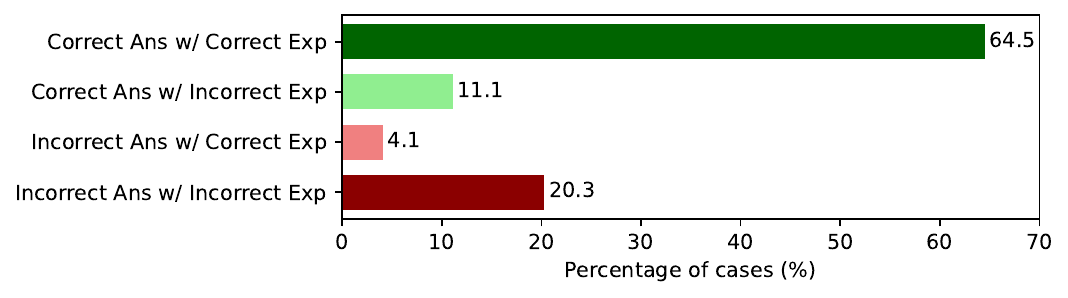}
        \caption{Chart QA}
        \label{subfig:1a}
    \end{subfigure}
    \hfill
    \begin{subfigure}{0.5\textwidth}
        \centering
        \includegraphics[width=\textwidth]{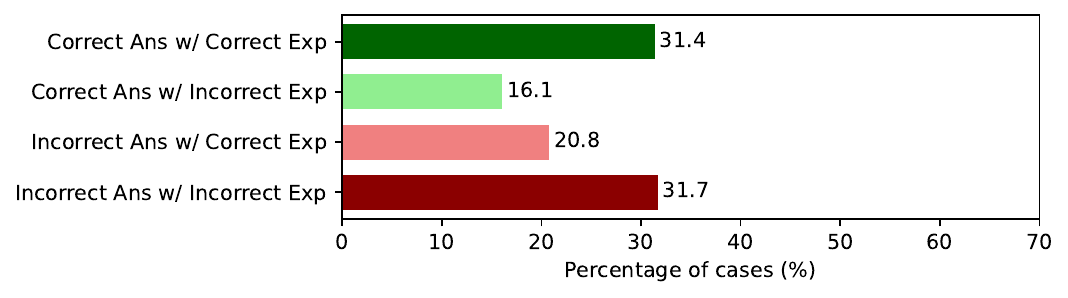}
        \caption{MathVista}
        \label{subfig:1b}
    \end{subfigure}

    \begin{subfigure}{0.49\textwidth}
        \centering
        \includegraphics[width=\textwidth]{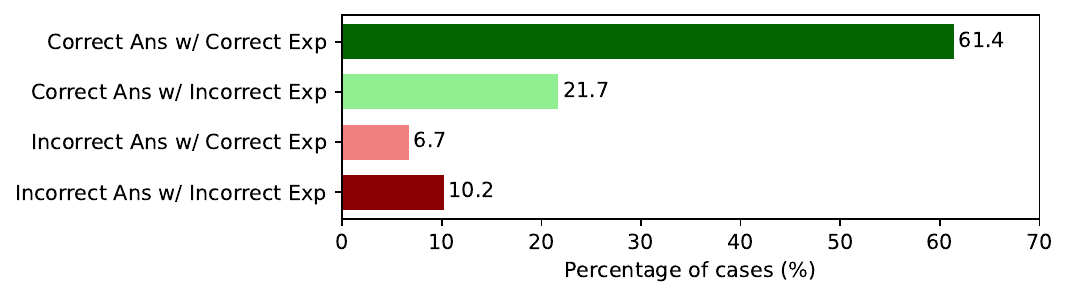}
        \caption{Spider}
        \label{subfig:2a}
    \end{subfigure}
    \hfill
    \begin{subfigure}{0.5\textwidth}
        \centering
        \includegraphics[width=\textwidth]{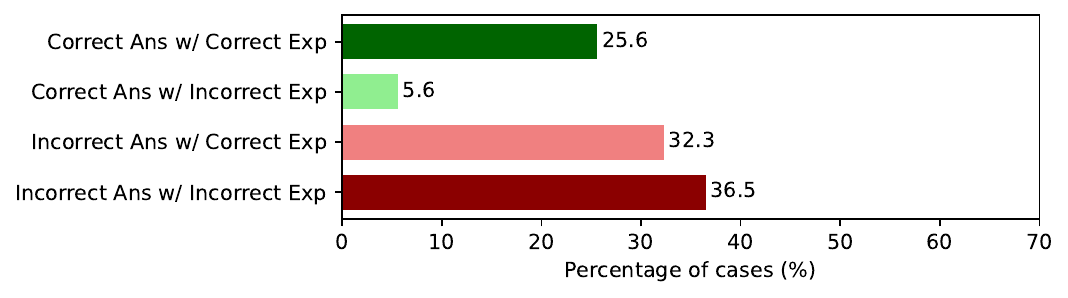}
        \caption{ARC}
        \label{subfig:2b}
    \end{subfigure}

    \caption{
Manual analysis of GPT-4V + VCoT 
on our sampled tasks.
We manually annotate the reasoning and the final answer separately for all benchmark dataset and present the analysis.
For Chart QA and Spider, we find that in 72 -- 85 \% cases the model generates the right explanation and is able to generate the correct answer from these. For MathVista and ARC that require more complex reasoning, we find that the model struggles to generate the right reasoning with only 51 -- 58 \% reasoning being correct.
    }
    \label{fig:detailed}
\end{figure*}

\subsubsection{Code Generation}

We evaluate NL-to-SQL generation in a multi-modal setting using Spider dataset \cite{yu-etal-2018-spider}.
Each task input consists of a relational database and natural language question pair, and the task output is the SQL code needed to answer that question.
To turn Spider tasks into multi-modal tasks, we consider only tasks over a single table and we render the first 50 rows into an image using \texttt{matplotlib.pyplot.table} \cite{matplotlib}.
We use the following instruction:

\texttt{Generate SQL query for the given user question. The relevant table is shown in the image and the metadata is included. \\
Metadata: <Metadata> \\
Question: <Question>
}

where we replace \texttt{<MetaData>} with 
the table schema (column names and types) and \texttt{<Question>} with the user question.
We use the schema encoding from FormaT5 \cite{singh2023format5}.
Note that we only evaluate GPT-4-based models on this problem, as other baseline methods have not been trained to generate code.

\begin{table}
\centering
\resizebox{\columnwidth}{!}{
\begin{tabular}{lllll} 
\toprule
\textbf{Approach}     & \textbf{MathVista} & \textbf{ChartQA} & \textbf{ARC}  & \textbf{Spider}  \\ \midrule
Sphinx       & 33.2      & 63.4    & 27.2 & -                            \\
InstructBlip & 24.7      & 54.5    & 23.4 & -                            \\
Blip2        & 23.1      & 31.4    & 22.6 & -                            \\
InstructBlip + CoT GPT-4    & 30.4      & 55.3    & 47.3 & 83.5                         \\
InstructBlip + PoT GPT-4    & 30.8      & 56.4    & \textbf{51.2} & 82.7                         \\
GPT-4        & -         & -       & -    & 81.8 \\
GPT-4V       & 47.5      & 75.6    & 31.2 & 84.3                            \\
GPT-4V + v-CoT       & \textbf{49.1}      & \textbf{79.2}    & 40.5 & \textbf{86.2}                            \\
\bottomrule
\end{tabular}
}
\caption{Results over 
20\% of tasks sampled from each benchmark uniformly at random. InstructBlip + GPT-4 indicates captioning by InstructBlip and reasoning over this by text GPT-4. CoT represents Chain of Thought; PoT represents Program of Thought. GPT-4 does not use a vision component and is only evaluated on problems that do not depend on an image.}
\label{table:summary-results}
\end{table}

\begin{figure*}[!t]
    \captionsetup{justification=centering}
    \centering
    \begin{subfigure}{0.33\textwidth}
        \centering
        \includegraphics[width=\textwidth]{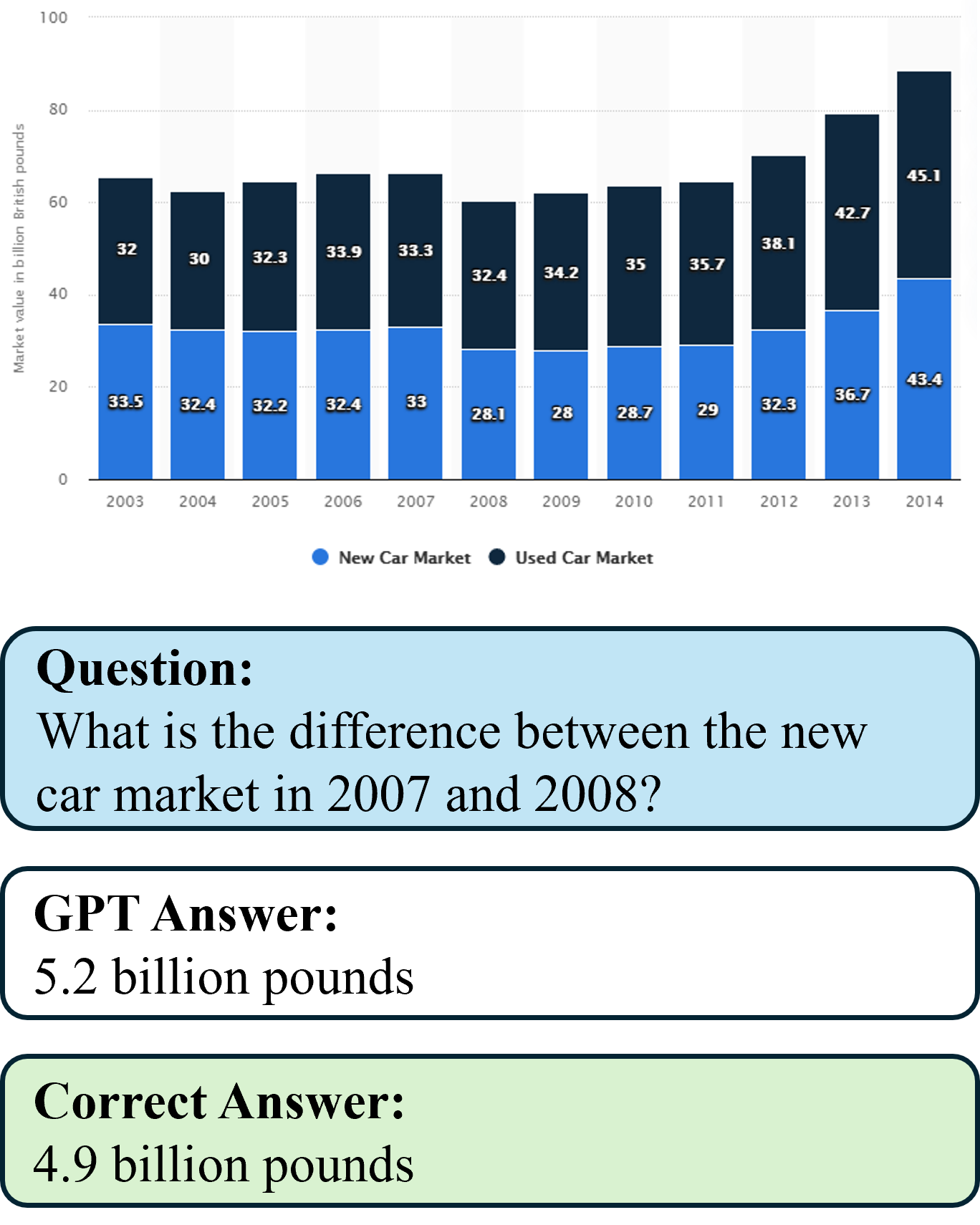}
        \caption{Computation Mistake}
        \label{subfig:chartqa-pattern-a}
    \end{subfigure}
    \hfill
    \begin{subfigure}{0.33\textwidth}
        \centering
        \includegraphics[width=\textwidth]{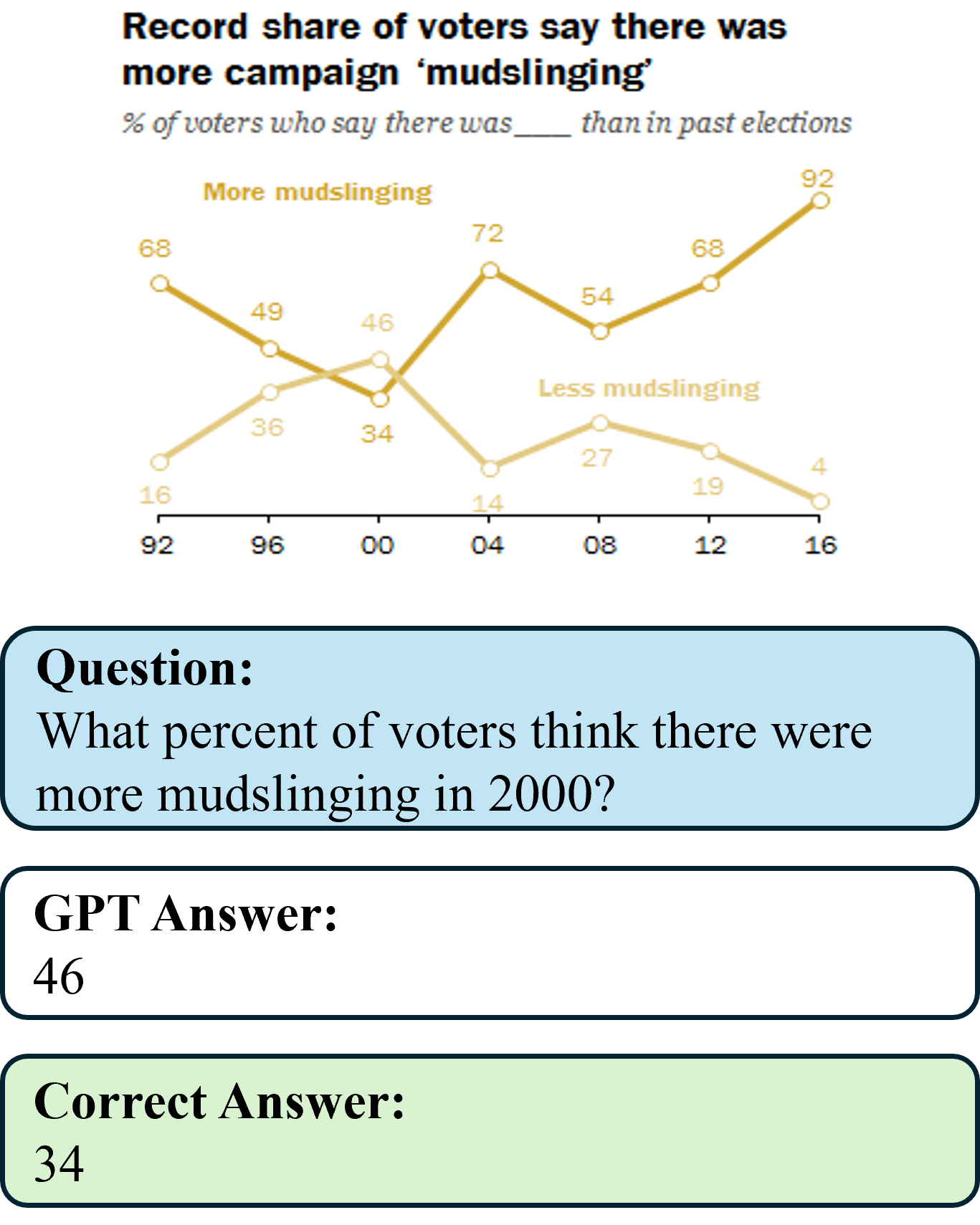}
        \caption{Color Inferring Mistake}
        \label{subfig:chartqa-pattern-b}
    \end{subfigure}
    \hfill
    \begin{subfigure}{0.33\textwidth}
        \centering
        \includegraphics[width=\textwidth]{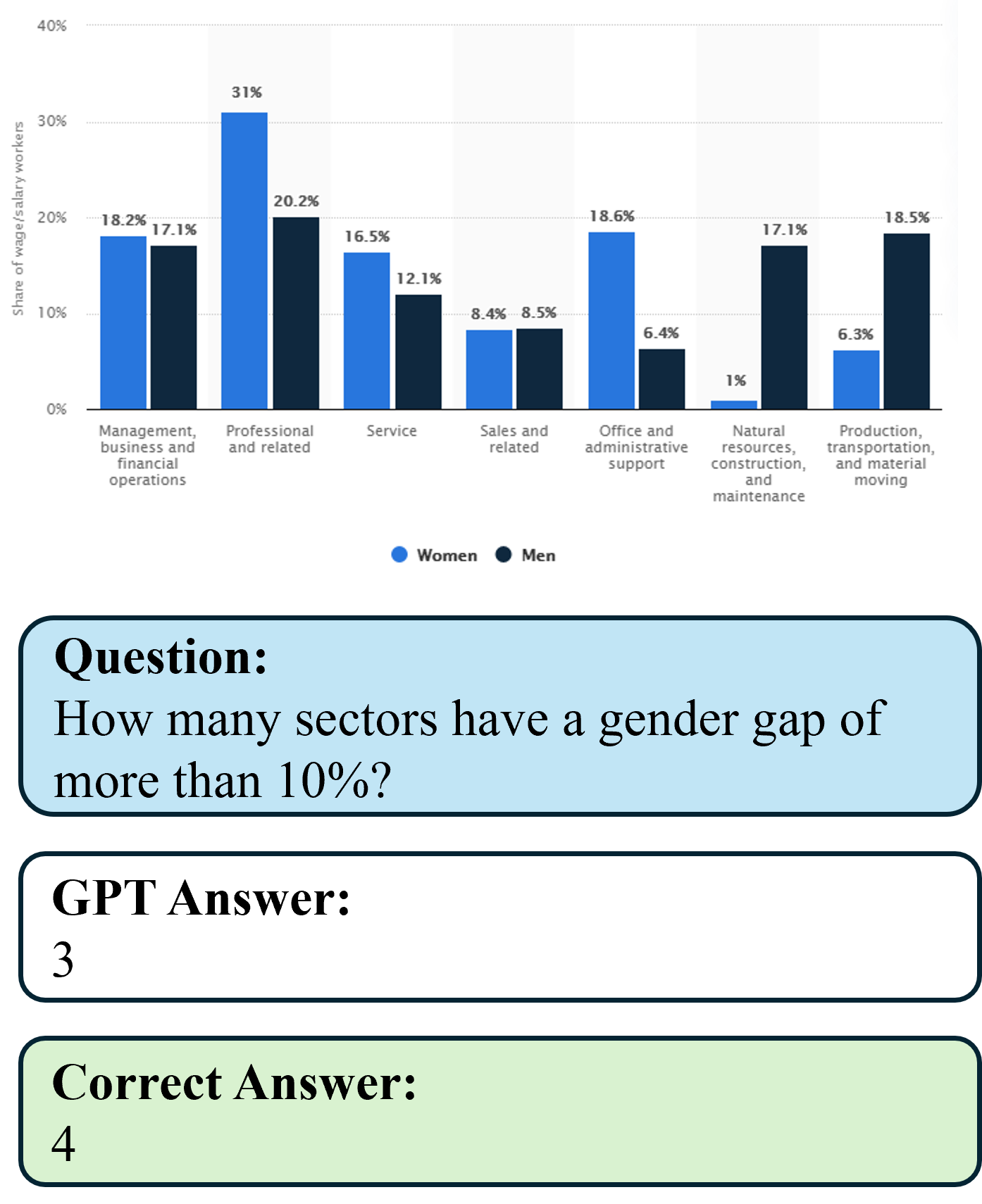}
        \caption{Text and Image Reasoning}
        \label{subfig:chartqa-pattern-c}
    \end{subfigure}

    \caption{
    Observed patterns in 
    GPT-4V behavior for
    ChartQA tasks.
    (a) shows an example where the model generates the correct reasoning, i.e. to subtract new car market share of 2008 from 2007, but makes a computation error in the final answer.
    (b) shows an example where the model generates the correct reasoning but applies it to the light yellow line instead of the dark yellow.
    (c) shows an example where the model needs to reason on both text and images together. V-CoT is able to fix this example.
    }
    \label{fig:chartqa-pattern-examples}
\end{figure*}

\begin{figure*}[!t]
    \captionsetup{justification=centering}
    \centering
    \begin{subfigure}{0.33\textwidth}
        \centering
        \includegraphics[width=\textwidth]{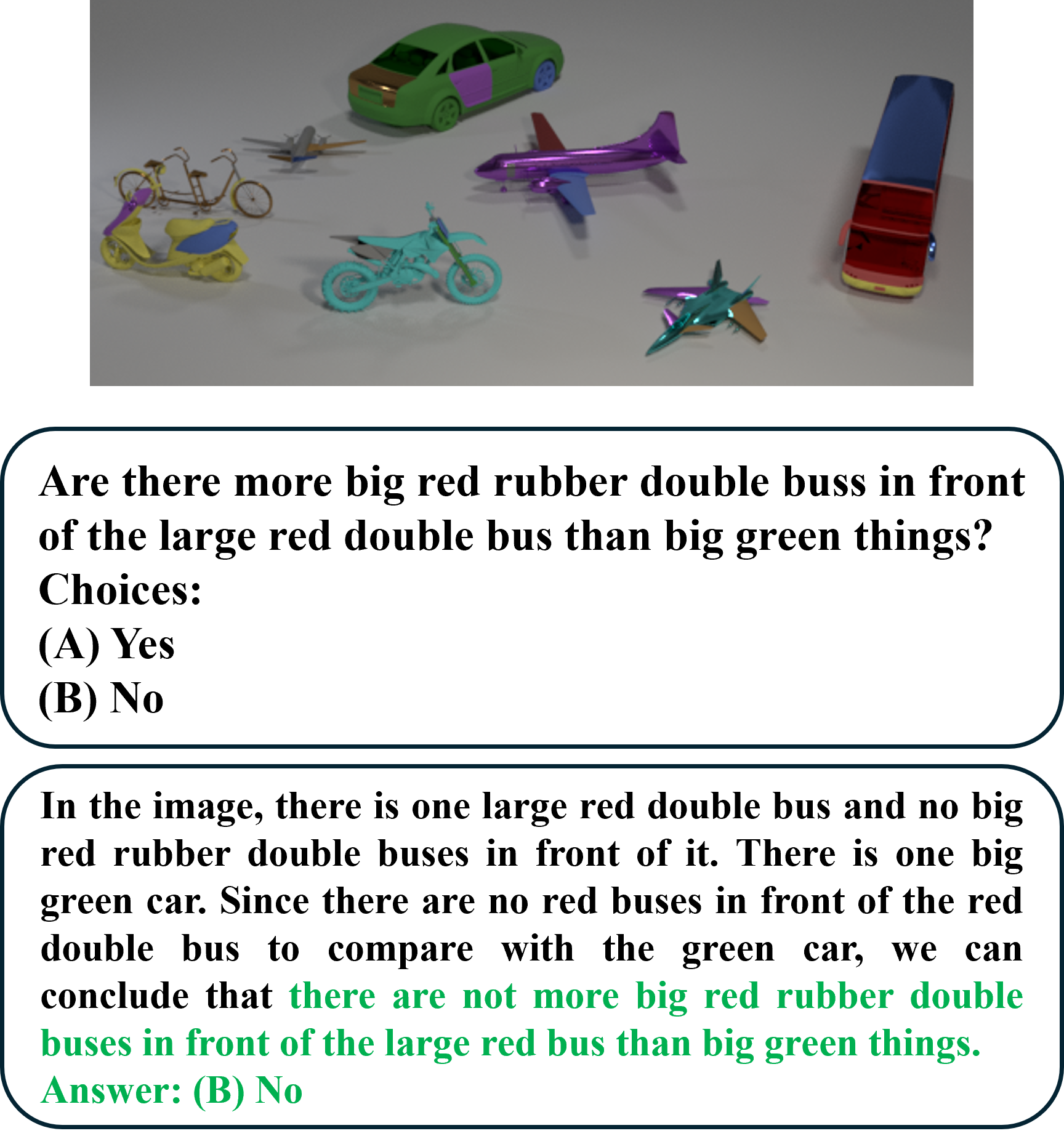}
        \caption{Correct Answer w/ Correct Reasoning}
        \label{subfig:mathvista-a}
    \end{subfigure}
    \hfill
    \begin{subfigure}{0.33\textwidth}
        \centering
        \includegraphics[width=\textwidth]{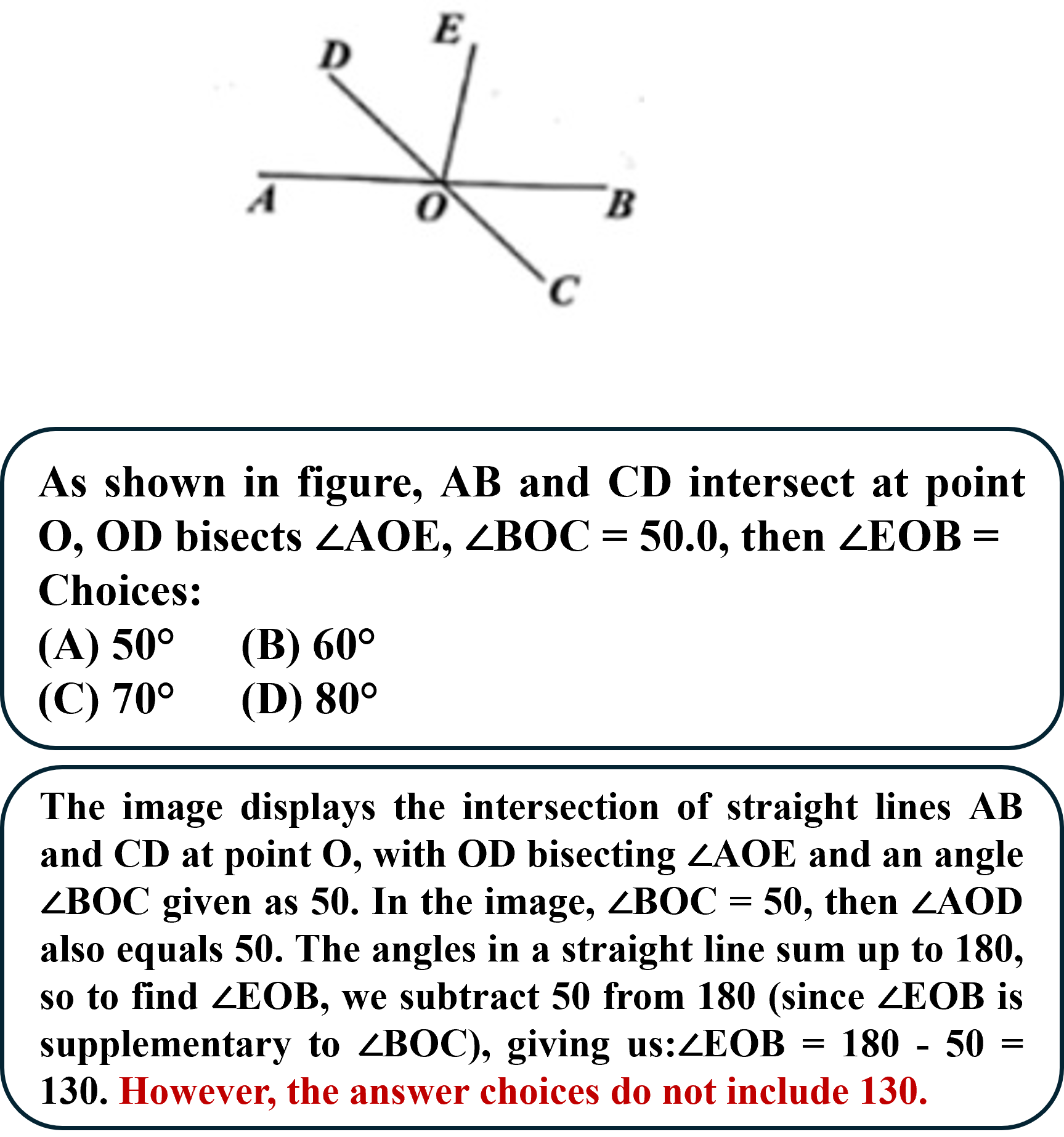}
        \caption{Incorrect Answer w/ Incorrect Reasoning}
        \label{subfig:mathvista-b}
    \end{subfigure}
    \hfill
    \begin{subfigure}{0.33\textwidth}
        \centering
        \includegraphics[width=\textwidth]{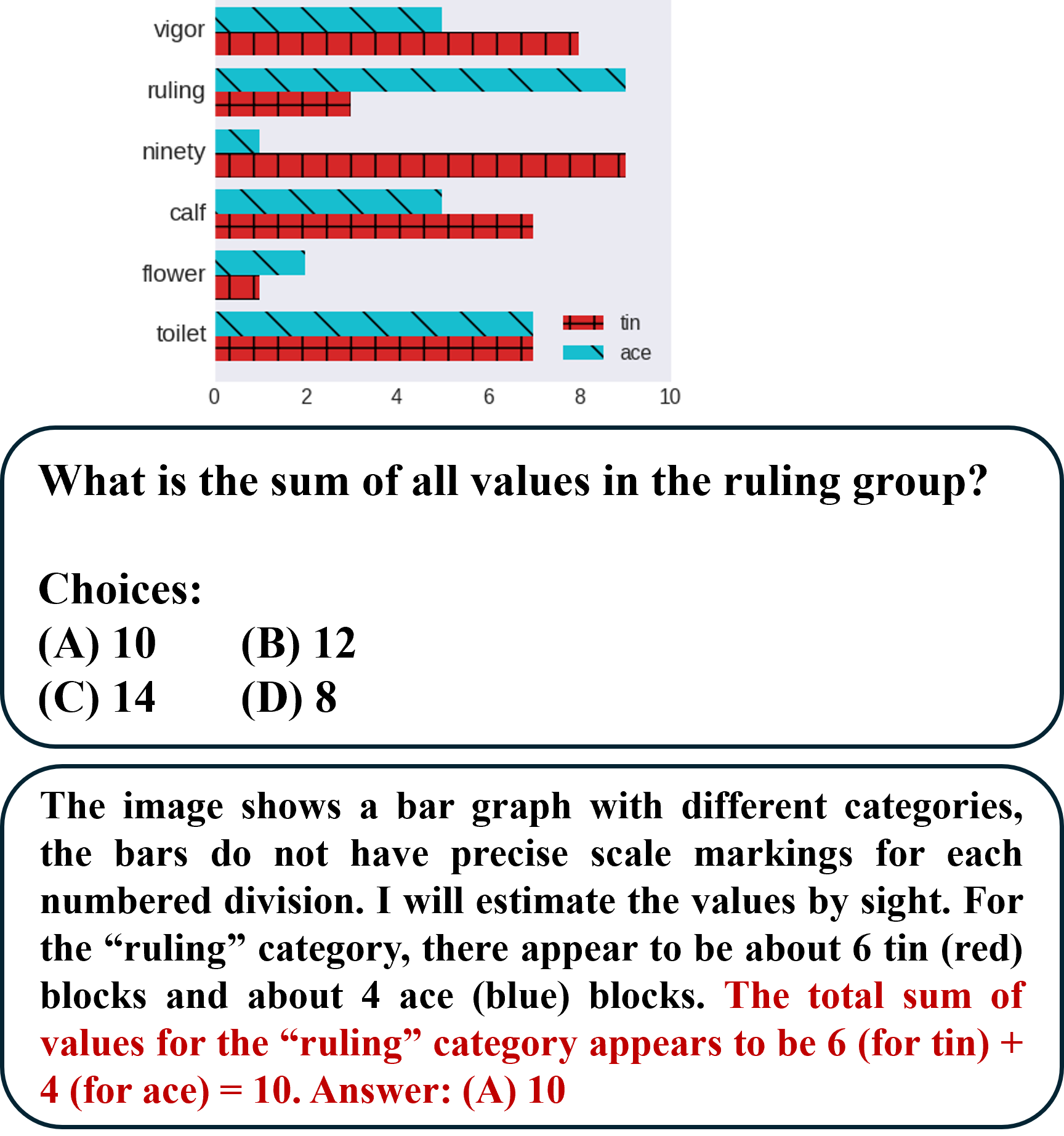}
        \caption{Incorrect Answer w/ Correct Reasoning}
        \label{subfig:mathvista-c}
    \end{subfigure}

    \caption{
    Observed patterns in
    GPT-4V + V-CoT responses
    to MathVista tasks.
    (a) shows an example where both the reasoning and answer are correct;
    (b) shows an example where both the reasoning and answer are incorrect as the model misses to subtract angle DOE;
    (c) shows an example where the answer is incorrect but the reasoning is correct, as the model correctly figures out the correct bar but cannot generate the correct value from scale.
    }
    \label{fig:mathvista-pattern-examples}
\end{figure*}

\section{Results}

Table~\ref{table:summary-results} presents a summary of our  results across benchmarks and baselines.
We find that for both MathVista and ChartQA, GPT-4V outperforms other approaches.
Interestingly, for the ARC dataset, we observe that using the multi-modal version of GPT-4 actually produces worse results, compared to generating textual descriptions and then applying GPT-4 plus CoT or PoT (the latter of which improves results most).
Overall, the ARC dataset represents a challenging task.
On the Spider dataset, we find that including image improves performance by 0.8 percentage points, while adding v-CoT improves by 1.9 percentage points.

We manually annotated GPT-4V results for whether the reasoning provided was correct or not and whether the answer was correct or not.
Figure~\ref{fig:detailed} presents our findings.
Across all datasets, we observe that there are tasks that GPT-4V correctly answers, despite having incorrect reasoning.
Similarly, we find that despite producing correct reasoning, the model can still produce an incorrect
answer.
This issue is particularly frequent in both the MathVista and ARC tasks.

\subsection{Discussion of Patterns}

We discuss recurring patterns we observed in our manual analysis of results.

\subsubsection{Mathematical Reasoning}

We identify the following recurring patterns in the MathVista failures.

\begin{itemize}
    \item Incorrect computation over derived values. Figure~\ref{subfig:mathvista-b} shows an example where the model identifies correct values and constraints for a trigonometric problem, but fails to compute the final answer.  
    \item Inferring numerical labels. When the model has to infer numerical labels, like aggregates in bar charts, the model often picks wrong values and fails to answer, despite having correct reasoning. This is shown in Figure~\ref{subfig:mathvista-c}.
\end{itemize}

\subsubsection{Visual Data Analysis}

We find that in ChartQA, multiple failures can be attributed to a few emergent behaviors, which are highlighted below.
These patterns are also illustrated in Figure~\ref{fig:chartqa-pattern-examples}.

\begin{itemize}
    \item Incorrect computation over chart-derived values. For example, take the question ``\emph{What is the growth from 2010 to 2011?}'' over the chart in Figure~\ref{subfig:chartqa-pattern-a}. Despite identifying each of these values correctly, it fails to compute the growth amount by subtracting the value of 2007 from that of 2008.
    \item Linking colors/visual elements. For example, when asked a question about a
    plot with two lines of varying shade of yellow, the model flips their labeling and
    answers with the wrong value (see 
    Figure~\ref{subfig:chartqa-pattern-b}).
    \item Computations that require extracting values from both image and the query. For example, extracting a rate from the text query and using this to identify differences in a plot (see Figure~\ref{subfig:chartqa-pattern-c}).
\end{itemize}


\subsubsection{Abstraction and Extrapolation}

In the ARC dataset, all methods struggle to generate the correct output since it requires complex reasoning over visual patterns.
We find the following to be a common pattern.

\begin{figure}[t]
    \centering
    \includegraphics[width=0.8\linewidth]{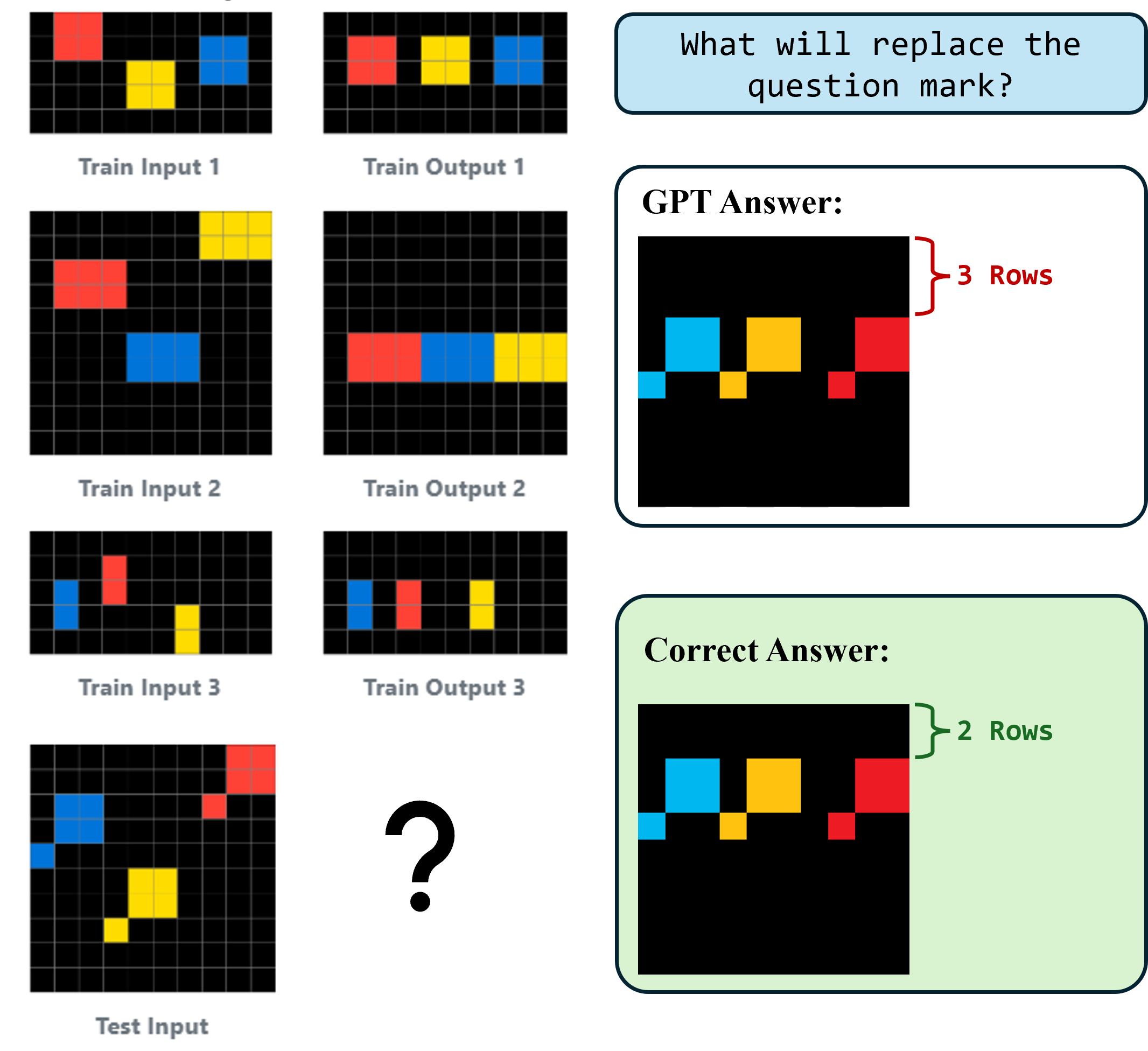}
    \caption{
In ARC, GPT-4V can still struggle
in identifying/placing blocks
in appropriate locations (like
offsets). The figure shows that GPT-4V correctly
identifies that it has to align the blocks to the level of the blue block,
but it is offset by one block.
    }
    \label{fig:arc-failure-patterns}
\end{figure}

\begin{itemize}
    \item Incorrect grid perception. Despite using v-CoT, we find cases where GPT-4V does not correctly
    identify block positions.
    For example, v-CoT may place blocks at the wrong vertical offset (see Figure~\ref{fig:arc-failure-patterns}).
\end{itemize}

\subsubsection{Code Generation}

\begin{figure}[t]
    \captionsetup{justification=centering}
    \centering
    \begin{subfigure}{0.475\linewidth}
        \centering
        \includegraphics[width=\linewidth]{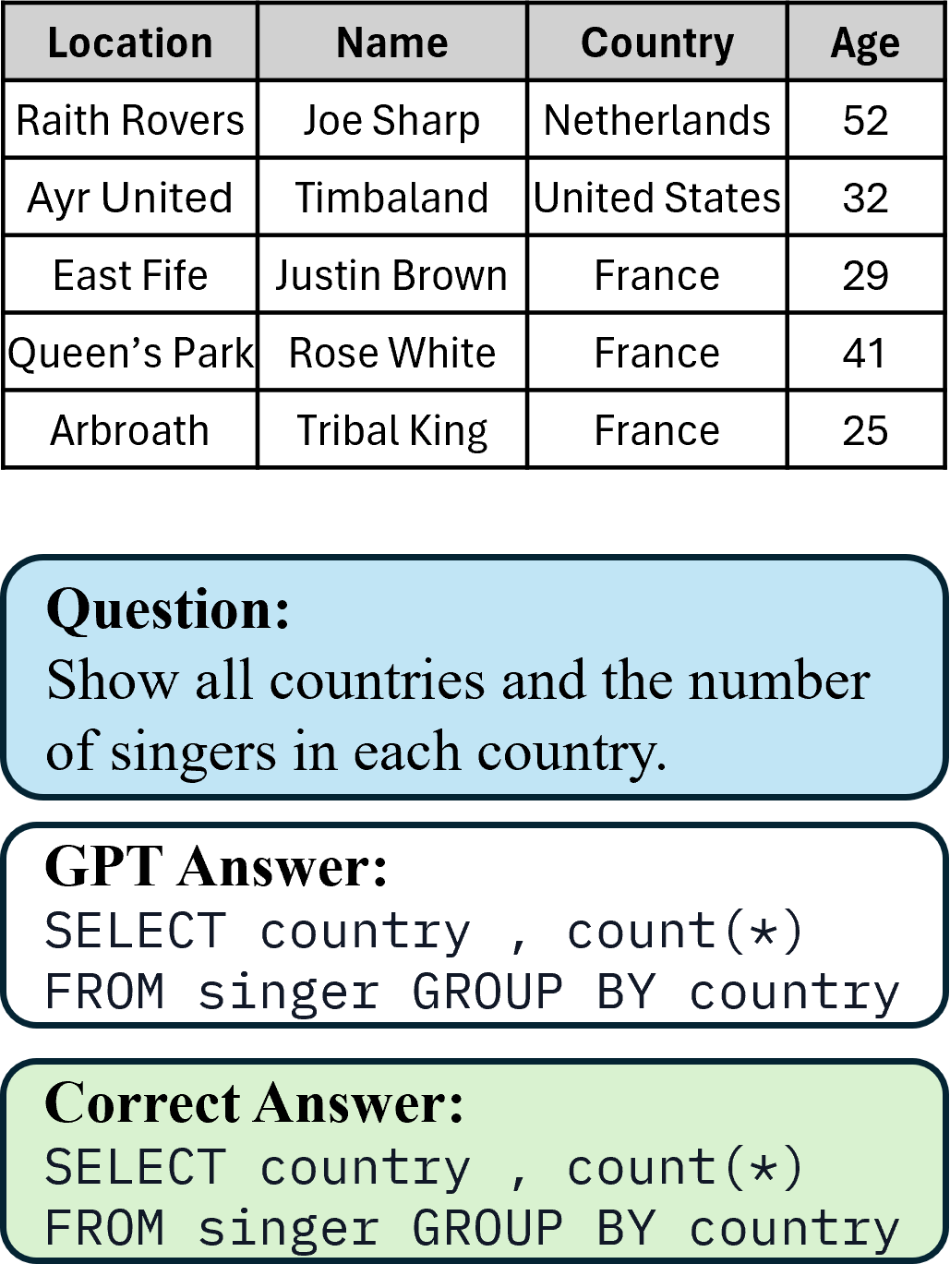}
        \caption{Success Case}
        \label{subfig:spider-failure-patterns-a}
    \end{subfigure}
    \hfill
    \begin{subfigure}{0.505\linewidth}
        \centering
        \includegraphics[width=\linewidth]{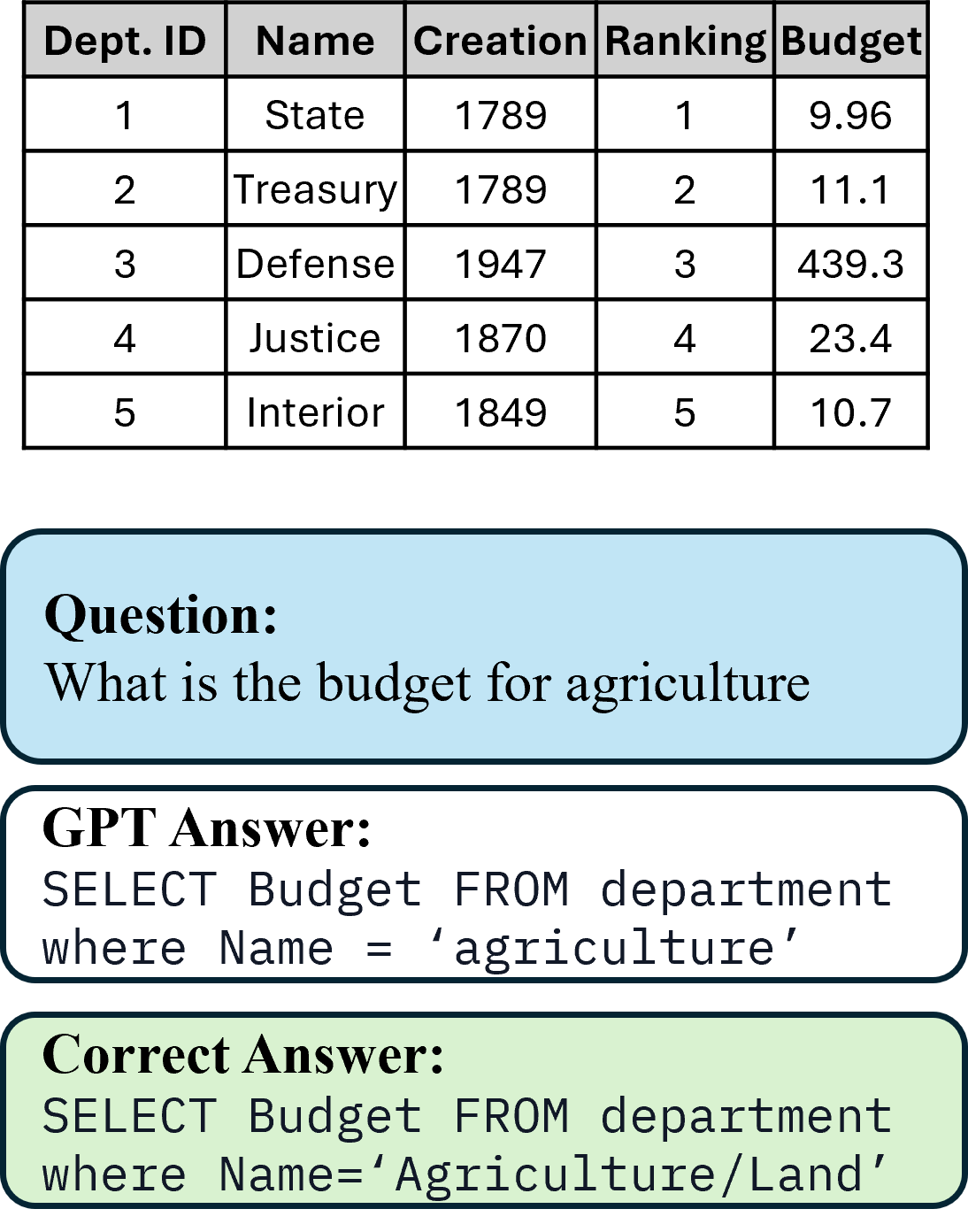}
        \caption{Failure Case}
        \label{subfig:spider-failure-patterns-b}
    \end{subfigure}

    \caption{We show success and failure cases of adding tables as images. (a) Adding the table image can help
for queries that requiring
overall table structure (like grouping) or
extracting particular values (like filtering) from table rendering.
(b) For queries that need constant values, GPT-4V struggles to get it from the image.
    }
    \label{fig:spider-failure-patterns}
\end{figure}

On SQL generation for Spider tasks, we make the following observations:

\begin{itemize}
    \item Structural understanding. The rendered table provides improvements over the text representation of the same table for tasks that require reasoning over the structure of a table (pivot, group, transposing). An example is shown in Figure~\ref{subfig:spider-failure-patterns-a}.
    \item Token efficiency of images. Considering tables as images consumed fewer tokens than their corresponding text description produced by InstructBlip.
    \item Reasoning over cell values. Tasks that require reasoning over cell values, such as filtering or splitting, benefit from a textual representation of tables or from the image paired with v-CoT (see Figure~\ref{subfig:spider-failure-patterns-b}).
\end{itemize}

\section{Conclusion}
We present the first evaluation of the state-of-the-art multi-modal LLM GPT-4V on structured reasoning tasks across various domains.
We show that our visual Chain-of-Thought (v-CoT) prompt improves performance by first instructing the model to analyse the image, conditioned on the task at hand, and then use this analysis to reason towards a final result.
Our experiments compare GPT-4V (with and without v-CoT) to multiple existing multi-modal models. 
We find that v-CoT outperforms other baselines in three datasets, but that it still struggles with the ARC
dataset, which requires abstraction and extrapolation.

\section{Limitations}

First, we consider a subset of structured reasoning tasks, performance on other tasks may vary.
Similarly, we sample 20\% of tasks per dataset to mitigate computational costs---while this sampling was done uniformly at random, it is possible that performance would change for other tasks.
While manually annotating results, we also identified that there are tasks in the various datasets (particularly ChartQA and MathVista) that are ambiguous or may have incorrect ground truth labels.
We take all labeling as given.
All our tasks are framed in English (images with English text, tasks with English text).
Evaluating GPT-4V and other models on multi-modal, multilingual tasks remains future work.

\nocite{langley00}

\bibliography{references}
\bibliographystyle{icml2023}



\end{document}